\documentclass[11pt,a4paper]{article}
\usepackage[hyperref]{acl2017}
\usepackage{times}
\usepackage{latexsym}
\usepackage{booktabs}
\usepackage{subcaption}
\usepackage{graphicx}
\usepackage{color,soul}
\usepackage{amsmath,amssymb}
\usepackage{multicol}
\usepackage{multirow}
\usepackage[normalem]{ulem}
\usepackage{pifont}
\newcommand{\cmark}{\ding{51}}%
\newcommand{\xmark}{\ding{55}}%
\usepackage{enumitem}
\usepackage{diagbox}
\usepackage{framed}

\usepackage{color}

\usepackage{url}

\aclfinalcopy 
\def\aclpaperid{496} 

\newcommand{\reals}{\mathbb{R}}

\newcommand{\sss}{\mathbf{s}} 

\title{What do Neural Machine Translation Models Learn about Morphology?}

\author{Yonatan Belinkov$^1$ ~~ Nadir Durrani$^2$ ~~ Fahim Dalvi$^2$ ~~ Hassan Sajjad$^2$ ~~ James Glass$^1$ \\\\
$^1$MIT Computer Science and Artificial Intelligence Laboratory, Cambridge, MA 02139, USA \\
{\tt \{belinkov, glass\}@mit.edu} \\
$^2$Qatar Computing Research Institute, HBKU, Doha, Qatar  \\
{\tt \{ndurrani, faimaduddin, hsajjad\}@qf.org.qa}
}

\date{}

\begin{document}
\maketitle
\begin{framed}
\noindent This is a modified version of a paper originally published at ACL 2017 with updated results and discussion in section 5. 
\end{framed}
\begin{abstract}
Neural machine translation (MT) models obtain state-of-the-art performance while maintaining a simple, end-to-end architecture. However, little is known about what these models learn about source and target languages during the training process. 
 In this work, we analyze the representations learned by neural MT models at various levels of granularity and empirically evaluate the quality of the 
 representations 
 for learning morphology 
 through extrinsic part-of-speech and morphological tagging tasks. We conduct a thorough investigation along several parameters: word-based vs.\ character-based representations, depth of the encoding layer, the identity 
 of the target language, and encoder vs.\ decoder representations. Our data-driven, quantitative evaluation sheds light on important aspects in the neural MT system and its ability to capture word structure.\footnote{Our code is available at \url{https://github.com/boknilev/nmt-repr-analysis}.}
\end{abstract}

\section{Introduction}

Neural network models are quickly becoming the predominant approach to machine translation (MT). Training neural MT (NMT) models can be done in an end-to-end fashion, which is simpler and more elegant than traditional MT systems. Moreover, NMT systems have become competitive with, or better than, the previous state-of-the-art, especially since the introduction of sequence-to-sequence models and the attention mechanism  \cite{bahdanau2014neural,sutskever2014sequence}. 
The improved translation quality is often attributed to better handling of non-local dependencies and morphology generation \cite{luong-manning:iwslt15,bentivogli-EtAl:2016:EMNLP2016,toral-sanchezcartagena:2017:EACLlong}. 
\bigskip

However, 
little is known about what and how much these models 
learn about each language and its features.
Recent work has started exploring the role of the NMT encoder in learning source syntax \cite{shi-padhi-knight:2016:EMNLP2016},
but 
research studies are yet to answer important questions such as: \textit{(i)} what do NMT models learn about word morphology? \textit{(ii)} what is the effect on learning when translating into/from morphologically-rich languages? \mbox{\textit{(iii)} what} impact do different representations (character vs.\ word) have  on learning? and \textit{(iv)} what do different  modules learn about the syntactic and semantic structure of a language? 
Answering such questions is imperative for fully understanding the NMT architecture. In this paper, we strive towards exploring \textit{(i)}, \textit{(ii)}, and \textit{(iii)} by providing quantitative, data-driven answers to the following specific questions:

\begin{itemize}

\item Which parts of the NMT architecture capture word structure? 

\item 
What is the division of labor between different components (e.g.\ different layers or 
encoder vs.\ decoder)? 

\item How do different word representations help learn better morphology and modeling of infrequent words?

\item How does the target language affect the learning of word structure? 
\end{itemize}

To achieve this, we follow a simple but effective procedure with three steps: \mbox{\textit{(i)} train} a neural MT system on a parallel corpus; \mbox{\textit{(ii)} use} the trained model to extract feature representations for words in a language of interest; and \mbox{\textit{(iii)} train} a classifier using extracted features to make predictions for another task. 
We then evaluate the quality of the trained classifier on the given task as a proxy to the quality of the extracted representations. In this way, we obtain a quantitative measure of how well the original MT system learns features that are relevant to the given task.  

We focus on the tasks of part-of-speech (POS) and full morphological tagging. We investigate how different neural MT systems capture POS and morphology through a series of experiments along several parameters. For instance, we contrast word-based and character-based representations, use different encoding layers, vary source and target languages, and compare extracting features from the encoder vs.\ the decoder.

We experiment with several languages with  varying degrees of morphological richness: 
French, German, Czech, Arabic, and Hebrew.\ 
Our analysis reveals interesting insights such as: 
\begin{itemize}
\item Character-based representations are much better for learning morphology, 
especially for low-frequency words. This improvement is correlated with better BLEU scores. On the other hand, word-based models are sufficient for learning the structure of common words.
\item Lower layers of the 
encoder are better at capturing word structure, while deeper networks improve translation quality, 
suggesting that higher layers focus more 
on word meaning. 
\item The target language impacts the kind of information learned by the MT system. Translating into morphologically-poorer languages leads to better source-side word representations. This is partly, but not completely, correlated with BLEU scores.
\item The NMT encoder and decoder learn representations of similar quality. The attention mechanism affects the quality of the encoder representations more than that of the decoder representations. 
\end{itemize}

\section{Methodology}

Given a source sentence $s = \{w_1, w_2, ..., w_N\}$ and a target sentence $t=\{u_1, u_2, ..., u_M\}$, we first 
generate a vector representation for the source sentence  
using 
an encoder (Eqn.\ \ref{eq:enc}) 
and then 
map this vector to the target sentence 
using 
a decoder (Eqn.\ \ref{eq:dec})
\cite{sutskever2014sequence}:
\begin{align} 
&\texttt{ENC}: s=\{w_1, w_2, ..., w_N\} \mapsto \sss \in \reals^k \label{eq:enc} \\
&\texttt{DEC} : \sss \in \reals^k \mapsto t=\{u_1, u_2, ..., u_M\}  \label{eq:dec}
\end{align}
In this work, 
we use long short-term memory (LSTM) \cite{hochreiter1997long} encoder-decoders with attention \cite{bahdanau2014neural},
which we train on parallel data.

\begin{figure}[t]
	\centering
	\includegraphics[width=\linewidth]{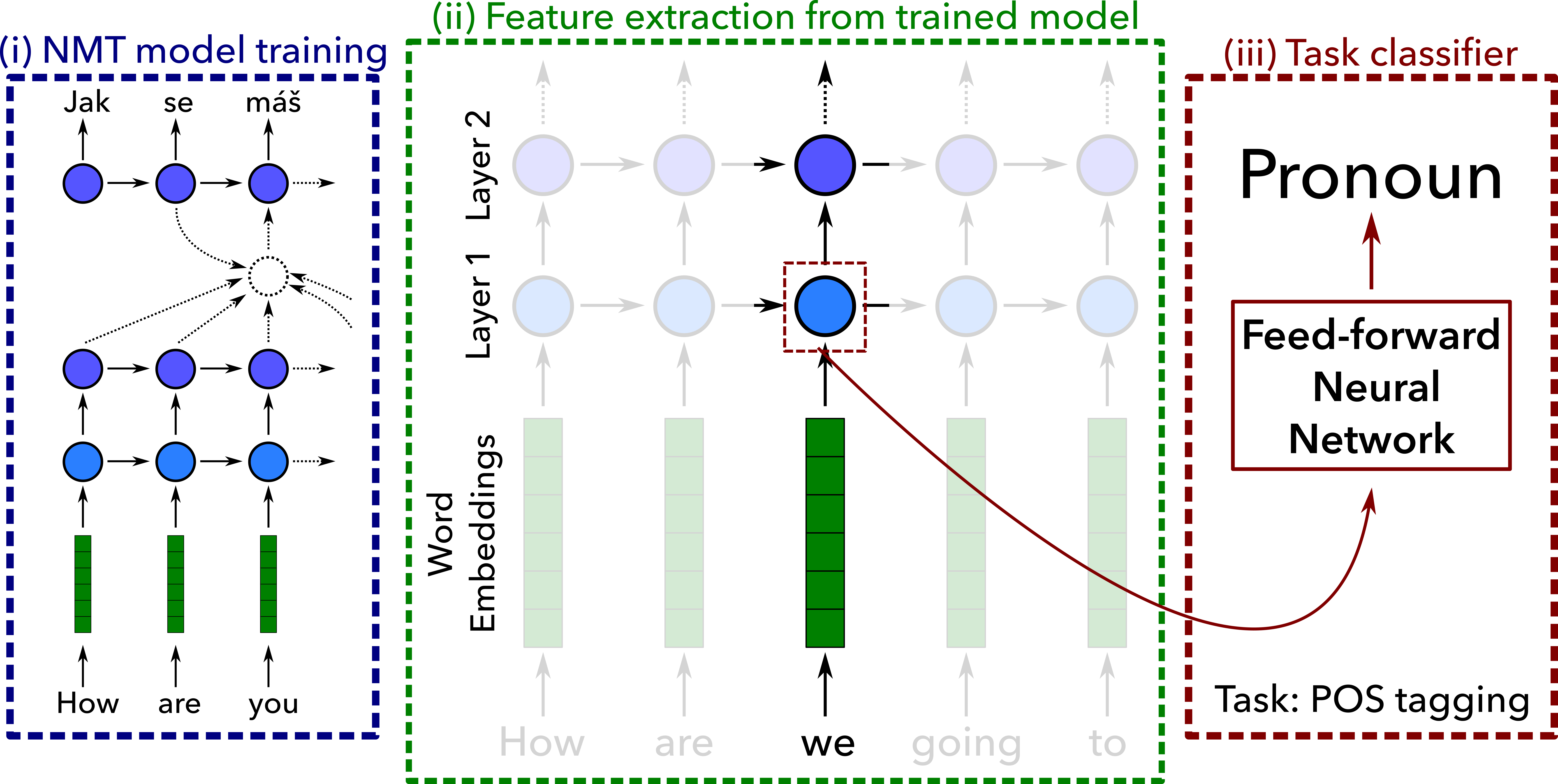}
	\caption{Illustration of 
    our approach: (i) NMT 
    system 
    trained on 
    parallel data; 
    (ii)  features extracted from pre-trained model;
    (iii) 
    classifier trained using the 
    extracted features. 
     Here a POS tagging classifier is trained on features from the first 
         hidden layer.
     }     
     	\label{fig:approach}
\end{figure}

After training the NMT system, we freeze the parameters of the encoder and use \texttt{ENC} as a feature extractor to generate vectors representing 
words in the sentence. Let $\texttt{ENC}_i(s)$ denote the encoded representation of word $w_i$. For example, this may be the output of the LSTM after word $w_i$. We 
feed $\texttt{ENC}_i(s)$ to a neural 
classifier that is trained to predict  POS or morphological tags and evaluate the quality of the representation based on our ability to train a good classifier. 
By comparing the performance of classifiers trained with features from different instantiations of \texttt{ENC}, we can evaluate what 
MT encoders learn about word structure.  Figure \ref{fig:approach} illustrates this process.
We follow a similar procedure 
for analyzing representation learning in $\texttt{DEC}$. 

The 
classifier itself can be modeled in different ways. For example, it may be an LSTM over outputs of the encoder. However, as we are interested in assessing the quality of the representations learned by the MT system, 
we choose to model the classifier as a simple feed-forward 
network with one hidden layer and a ReLU non-linearity. Arguably, if the learned representations are good, then a non-linear classifier should be able to extract useful information from them.\footnote{We also experimented with  a linear classifier and observed similar trends to the non-linear case, but overall lower results; \newcite{qian-qiu-huang:2016:P16-11} reported similar findings.}
We emphasize that our goal is not to beat the state-of-the-art on a given task, but rather to analyze what NMT  models %
learn about morphology.
The classifier is trained with a cross-entropy loss;
more details on 
its architecture are 
in the supplementary material.

\section{Data}

\begin{table}[t]
\centering
\footnotesize
\begin{tabular}{l|r|r|r|r}
\toprule
& \multicolumn{1}{c}{Ar} & \multicolumn{1}{c}{De} & \multicolumn{1}{c}{Fr} & \multicolumn{1}{c}{Cz} \\
\midrule
& \multicolumn{1}{c}{Gold/Pred} & \multicolumn{1}{c}{Gold/Pred} & \multicolumn{1}{c}{Pred} & \multicolumn{1}{c}{Pred} \\
\midrule
Train Tokens & 0.5M/2.7M & 0.9M/4.0M & 5.2M & 2.0M \\
Dev Tokens & 63K/114K & 45K/50K & 55K & 35K\\
Test Tokens & 62K/16K & 44K/25K & 23K & 20K\\
\midrule
POS Tags & 42 & 54 & 33 & 368 \\
Morph Tags & 1969 & 214 & -- & -- \\
\bottomrule
\end{tabular}
\caption{Statistics for annotated corpora in Arabic
(Ar), German (De), French (Fr), and Czech (Cz).}
\label{tab:tagsets}
\end{table}

\paragraph{Language pairs
} We experiment with several language pairs, including morphologically-rich languages, that have received relatively significant attention in the MT community. These include Arabic-, German-, French-, and Czech-English pairs. To broaden 
our analysis
and study the effect of having morphologically-rich languages on both source and target sides, we also  
include 
Arabic-Hebrew, two languages with rich and similar morphological systems, and Arabic-German, two languages with rich but different morphologies.

\paragraph{MT 
data}
Our translation 
models are trained on the WIT$^3$ corpus of TED talks \cite{cettoloEtAl:EAMT2012,cettolol:SeMaT:2016} made available for IWSLT 2016. This allows for comparable and cross-linguistic analysis. Statistics about each language pair are given in Table \ref{tab:tagsets} (under Pred). 
We use official dev and test sets for tuning and testing. Reported figures are the averages over test sets. 

\paragraph{Annotated data} 
We use two kinds of datasets to train POS and morphological classifiers: gold-standard and predicted tags. For predicted tags, we simply used freely available taggers 
to annotate the MT data. For gold tags, we use gold-annotated datasets. Table \ref{tab:tagsets} gives 
statistics for datasets with gold and predicted tags; see 
supplementary material
for 
details on 
taggers and gold data. 
We train and test our classifiers on predicted annotations, and similarly on gold annotations, when we have them. We report both results wherever available.

\section{Encoder Analysis} \label{sec:enc-analysis}

\begin{table}[t]
\centering
\begin{tabular}{l|c|c|r}
\toprule
& \multicolumn{1}{c}{Gold} & \multicolumn{1}{c}{Pred} & \multicolumn{1}{c}{BLEU} \\
\midrule
& \multicolumn{1}{c}{Word/Char} & \multicolumn{1}{c}{Word/Char} & \multicolumn{1}{c}{Word/Char} \\
\midrule
Ar-En & 80.31/93.66 & 89.62/95.35 & 24.7/28.4 \\
Ar-He & 78.20/92.48 & 88.33/94.66 & 9.9/10.7  \\
De-En & 87.68/94.57 & 93.54/94.63 & 29.6/30.4 \\
Fr-En & -- & 94.61/95.55 & 37.8/38.8 \\
Cz-En & -- & 75.71/79.10 & 23.2/25.4 \\
\bottomrule
\end{tabular}
\caption{POS accuracy on gold and predicted tags using word-based and character-based representations, as well as corresponding BLEU scores.
}
\label{tab:results-all-pairs}
\end{table}

Recall that after training the NMT system we freeze its parameters and use it only to generate features for the POS/morphology classifier.  Given a trained encoder \texttt{ENC} and a sentence $s$ with POS/morphology annotation, we generate word features $\texttt{ENC}_i(s)$ for every word in the sentence. We then train a classifier that uses the features $\texttt{ENC}_i(s)$ to predict POS or morphological tags.

\subsection{Effect of word representation}

\begin{figure*}[t]
    \centering
    \begin{subfigure}[t]{0.45\textwidth}
        \includegraphics[width=\textwidth]{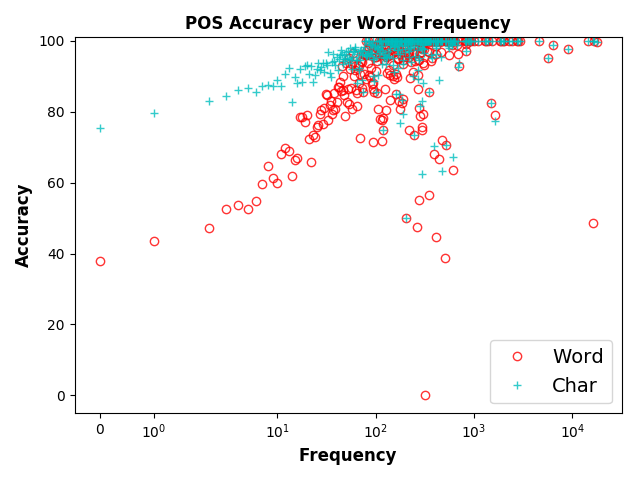}
    \end{subfigure}\hfill
    \begin{subfigure}[t]{0.45\textwidth}
       \includegraphics[width=\textwidth]{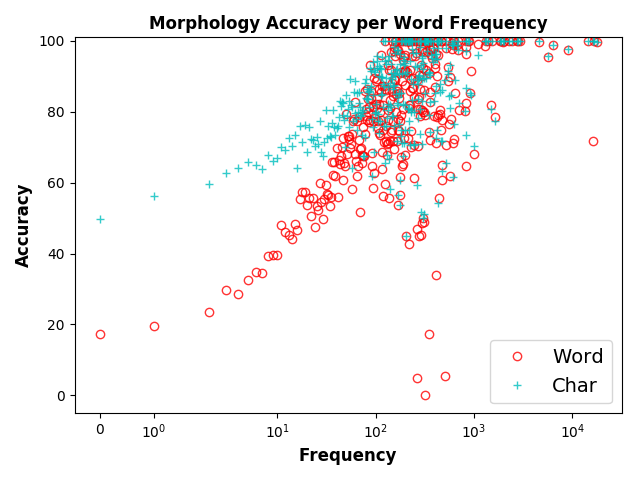}
    \end{subfigure}
      \caption{      POS and morphological tagging accuracy of word-based and character-based models per word frequency in the training data. Best viewed in color. 
      }
      \label{fig:repr-freqs}
\end{figure*}

\begin{figure}[t]
\includegraphics[width=\linewidth]{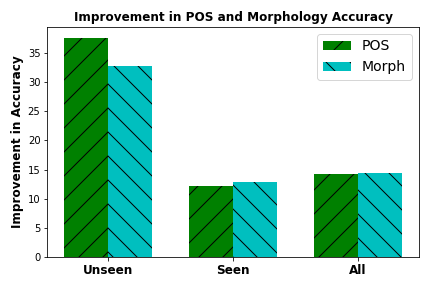}
\caption{Improvement in POS/morphology accuracy of character-based vs. 
word-based models 
for words unseen/seen in training, and for all words.}
\label{fig:repr}
\end{figure}

In this section, we compare different word representations extracted with different encoders. Our word-based model uses a word embedding matrix which is initialized randomly and learned with other NMT parameters. For a character-based model we adopt a convolutional neural network (CNN) over character embeddings that is also learned during training \cite{kim2015character,costajussa-fonollosa:2016:P16-2}; 
see appendix \ref{sec:sup-training} for specific settings. 
In both cases we run the encoder over these representations and use its output $\texttt{ENC}_i(s)$ as features for the classifier. 

Table~\ref{tab:results-all-pairs} shows 
POS tagging accuracy
using features from different NMT encoders. Char-based models always generate better representations for POS tagging, especially 
in the case of morphologically-richer languages like Arabic and Czech.  
We observed a similar pattern in the full morphological tagging task. For example, we obtain morphological tagging accuracy of 65.2/79.66 and  67.66/81.66 using word/char-based representations from the Arabic-Hebrew and Arabic-English encoders, respectively.\footnote{The results are not  far below 
dedicated taggers  (e.g.\ 95.1/84.1 on Arabic POS/morphology 
\cite{PASHA14.593.L14-1479}), indicating 
that NMT models 
learn quite 
good representations.} 
The superior morphological power of the char-based model also manifests in better translation quality (measured by BLEU), as shown in  Table~\ref{tab:results-all-pairs}.

\paragraph{Impact of word frequency}
Let us look more closely at an example case: Arabic POS and morphological tagging. 
Figure~\ref{fig:repr} shows the effect of using word-based vs.\ char-based feature representations, obtained from the encoder of the Arabic-Hebrew system (other language pairs exhibit similar trends).  Clearly, the char-based model is superior to the word-based one. This is 
true for the overall accuracy (+14.3\% in POS, +14.5\% in morphology), but 
more so on 
OOV words (+37.6\% in POS, +32.7\% in morphology). 
Figure~\ref{fig:repr-freqs} shows that the gap between word-based and char-based representations increases as the frequency of the word in the training data decreases. In other words, the more frequent the word, the less need there is for character information. These findings make intuitive sense: the char-based model is able to learn character n-gram patterns that are important for identifying word structure, but as the word becomes more frequent the word-based model has seen enough examples 
to make a decision.

\paragraph{Analyzing specific tags}

\begin{figure}[t]
\includegraphics[width=\linewidth]{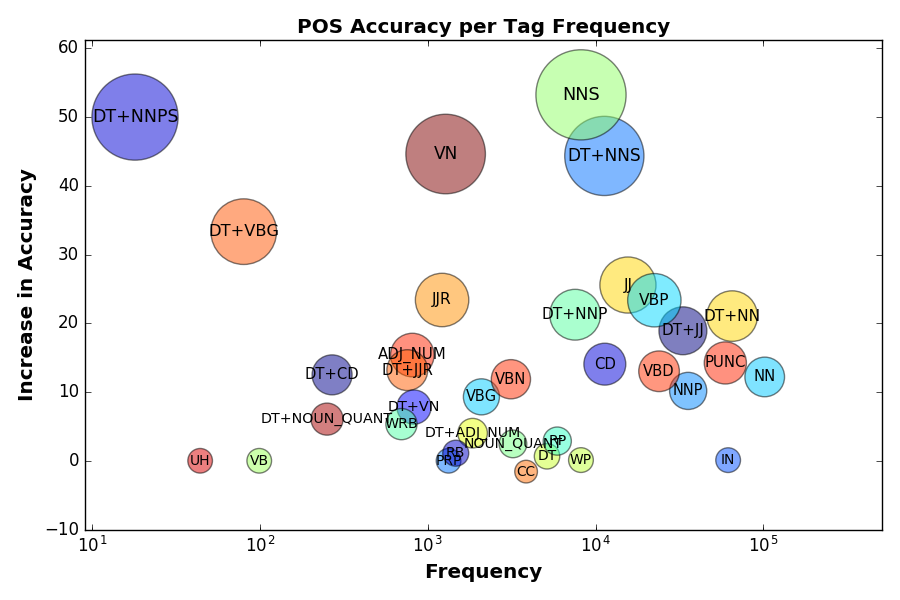}
\caption{Increase in POS accuracy with char- vs.\ word-based representations per tag frequency in the training set; larger bubbles reflect greater gaps.}
\label{fig:repr-pos-tag-freq}
\end{figure}

\begin{figure*}[t]
    \centering
    \begin{subfigure}[t]{0.49\textwidth}
        \includegraphics[width=\textwidth]{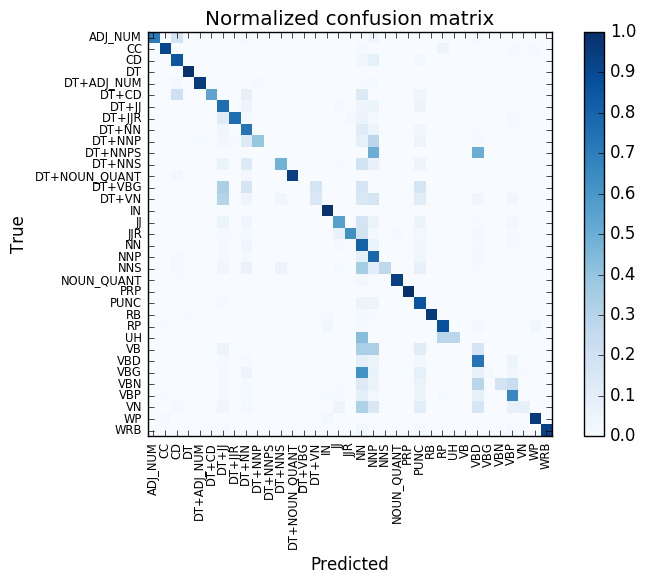}
        \caption{Word-based representations.}
        \label{fig:repr-pos-word-cm}
    \end{subfigure}\hfill
    \begin{subfigure}[t]{0.49\textwidth}
       \includegraphics[width=\textwidth]{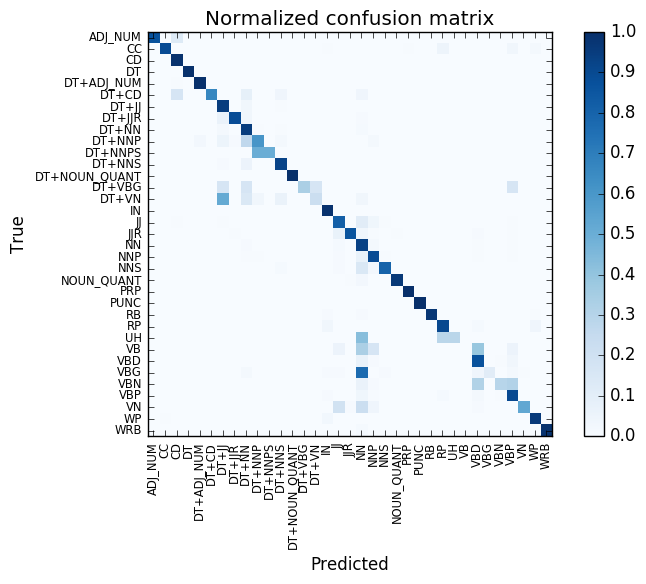}
        \caption{Character-based representations.}
        \label{fig:repr-pos-char-cm}
    \end{subfigure}
      \caption{Confusion matrices for POS tagging using word-based and character-based representations.}
      \label{fig:repr-pos-cm}
\end{figure*}

In Figure~\ref{fig:repr-pos-cm} we plot confusion matrices for POS tagging using word-based and char-based representations (from Arabic encoders).
While the char-based representations are overall better, the two models still share similar misclassified tags. 
Much of the confusion comes from wrongly predicting nouns (NN, NNP). In the word-based case, 
relatively many 
tags with determiner (DT+NNP, DT+NNPS, DT+NNS, DT+VBG) are wrongly predicted as non-determined nouns (NN, NNP). In the char-based case, this hardly  
happens. This suggests that 
char-based representations 
are predictive of the presence of a determiner, which in Arabic is expressed as the prefix ``Al-'' (the definite article), a pattern easily captured by a char-based model.

In Figure~\ref{fig:repr-pos-tag-freq} we plot the difference in POS accuracy when moving from word-based to char-based representations, per POS tag frequency in the training data. Tags closer to the upper-right corner 
occur more frequently in the training set and are better predicted by  char-based compared to word-based representations. 
There are a few 
fairly frequent tags (in the middle-bottom part of the figure) whose accuracy does not improve much when moving from word- to char-based representations: 
mostly conjunctions, determiners, and certain particles (CC, DT, WP). But there are several very frequent tags (NN, DT+NN, DT+JJ, VBP, and even PUNC) whose accuracy improves quite a lot. Then there are plural nouns (NNS, DT+NNS) where the char-based model really shines, which makes sense linguistically as plurality in Arabic is usually expressed by certain suffixes (\mbox{``-wn/yn''} for masc. plural, \mbox{``-At''} for fem. plural).  
The char-based model is thus especially good with frequent tags and infrequent words, which 
is understandable given that infrequent words typically belong to frequent open categories like nouns and verbs.

\subsection{Effect of encoder depth}

\begin{figure}[t]
\includegraphics[width=\linewidth]{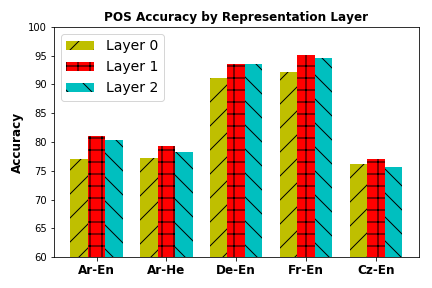}
\caption{POS tagging accuracy using representations from layers 0 (word vectors), 1, and 2, taken from encoders of different language pairs.}
\label{fig:layer-effect-all-langs}
\end{figure}

Modern NMT systems use very deep architectures with up to 8 or 16 layers \cite{wu2016google,TACL863}. We would like to understand 
what kind of information different layers capture. Given a trained 
model with multiple layers, we extract  
representations from the 
different layers in the encoder. Let $\texttt{ENC}^l_i(s)$ denote the encoded representation of word $w_i$ after the $l$-th layer. We  
vary $l$ and train different classifiers to predict POS or morphological tags.  Here we focus on the case of a 2-layer encoder-decoder 
for simplicity ($l \in \{1,2\}$).

\begin{figure}[t]
\includegraphics[width=\linewidth]{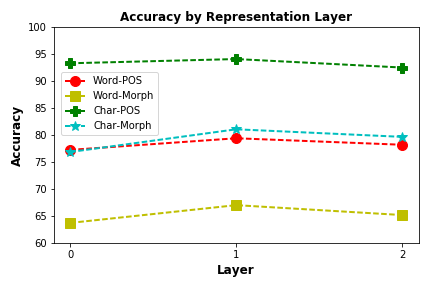}
\caption{POS and morphological tagging accuracy 
across layers. Layer 0: word vectors or char-based representations before the encoder; layers 1 and 2: representations after the 1st and 2nd layers.}
\label{fig:layer-effect-lines}
\end{figure}

Figure~\ref{fig:layer-effect-all-langs} shows POS tagging results using representations from different encoding layers across five 
language pairs. The general trend is that passing word vectors through the
encoder improves POS 
tagging, which can be explained by 
contextual information contained in the representations after one layer. However, 
it turns out that representations from the 1st layer are better than those from the 2nd layer, at least for the purpose of capturing word structure.
Figure~\ref{fig:layer-effect-lines} shows 
that the same pattern holds for both word-based and char-based representations, on Arabic POS and morphological tagging. In all cases, layer 1 representations are better than layer 2 representations.\footnote{We found this result to be also true in French, German, and Czech experiments  
(see
the supplementary material).
}  
In contrast, BLEU scores actually increase when training \mbox{2-layer} vs.\ \mbox{1-layer} models (+1.11/+0.56 BLEU for Arabic-Hebrew word/char-based models). 
Thus translation quality improves when adding layers but morphology quality degrades. 
Intuitively, it seems that lower layers of the network learn to represent word structure
while higher layers focus more 
on word meaning. 
A similar pattern was recently observed in a joint language-vision deep recurrent net~\cite{gelderloos-chrupala:2016:COLING}.

\begin{figure}[t]
\includegraphics[width=\linewidth]{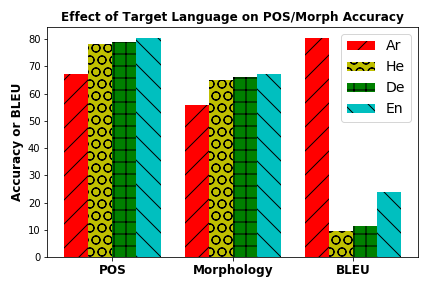}
\caption{Effect of target language on representation quality of the Arabic source.}
\label{fig:target-lang}
\end{figure}

\subsection{Effect of target language}

While translating from morphologically-rich languages is  
challenging, 
translating into such languages is even harder. 
For instance, our basic system obtains BLEU 
of 24.69/23.2 on Arabic/Czech to English, but only 13.37/13.9 on English to Arabic/Czech. 
How does the target language affect
the learned source language representations? Does translating into a morphologically-rich language require more knowledge about source language morphology? In order to investigate these questions, we fix the source language and train NMT models on 
different target languages. 
For example, given an Arabic source 
we train Arabic-to-English/Hebrew/German 
systems. 
These target languages represent a morphologically-poor language (English), a morphologically-rich language with similar morphology to the source language (Hebrew), and a morphologically-rich language with different morphology (German).  
To make a fair comparison, we train the models on the intersection of the training data based on the source language. In this way the experimental setup is completely identical: the models are trained on the same Arabic sentences with different translations.

Figure~\ref{fig:target-lang} shows POS and morphology accuracy of word-based representations from the NMT encoders, as well as  
corresponding BLEU scores. As expected, translating to 
English is easier than translating to 
the morphologically-richer Hebrew and German, resulting in higher BLEU. 
Despite their similar morphologies, 
translating Arabic to Hebrew is worse than Arabic to German, which can be attributed to the 
richer Hebrew morphology 
compared to German. POS and morphology accuracies share an intriguing pattern: the representations that are learned when translating to  
English are better for predicting POS or morphology than those learned when translating to 
German, which are in turn better than those learned when translating to 
Hebrew. This is remarkable given that English is a morphologically-poor language that  
does not display many of the 
morphological properties that are found in the Arabic source. In contrast, German and Hebrew have richer morphologies, so one could expect that translating into them 
would make the model learn more about morphology.

A possible explanation for this phenomenon is that the Arabic-English model is simply better 
than the Arabic-Hebrew and Arabic-German models, as hinted by the BLEU scores in Table \ref{tab:results-all-pairs}. The inherent  
difficulty in translating Arabic to Hebrew/German  
may affect the ability to learn good representations of word structure. 
To probe this more, we 
trained an Arabic-Arabic autoencoder 
on the same training data. We found that it learns 
to recreate the test sentences extremely well, with very high BLEU scores (Figure~\ref{fig:target-lang}). However, its word representations are actually inferior for the purpose of POS/morphological tagging. This 
implies that higher BLEU does 
not necessarily entail better morphological representations. In other words, a better translation model learns more informative representations, but only when it is actually learning to translate rather than merely memorizing the data as in the autoencoder case. 
We found 
this to be consistently true also for char-based experiments, 
and in other language pairs.

\section{Decoder Analysis} \label{sec:dec-analysis}

So far we only looked at the encoder. However, the decoder \texttt{DEC} 
is a crucial part in an MT system with access to both source and target sentences. In order to examine what the decoder learns about morphology, we first train an NMT system on the parallel corpus. Then, we use the trained model to encode a source sentence and extract features for words in the target sentence. 
These features are used to train a classifier on POS or morphological tagging on the target side.\footnote{In this section we only experiment with 
predicted tags 
as there are no 
parallel data 
with gold POS/morphological 
tags 
that we are aware of.} %
Note that in this case the decoder is given the correct target words one-by-one, similar to the 
usual NMT training regime.

Table \ref{tab:pos-dec-enc-attn-nogold} (1st row) shows the results of using 
representations extracted with \texttt{ENC} 
and \texttt{DEC} 
from the Arabic-English and English-Arabic models, respectively. 
There is a modest drop in representation quality with the decoder. This drop may be correlated with lower BLEU scores when translating English to Arabic vs.\ Arabic to English. We observed simmilar small drops with higher quality translation directions (Table~\ref{tab:decoder}, Appendix~\ref{sec:sup-results}).

The little gap between encoder and decoder representations may sound surprising, when we consider the fundamental tasks of the two modules. The encoder's task is to create a generic, close to language-independent representation of the source sentence, as shown by recent evidence from multilingual NMT \cite{johnson2016google}. The decoder's task is 
to use this representation to generate the 
target sentence in a specific language. 
One might conjecture that it would be sufficient for the decoder to learn a strong language model in order %
to produce morphologically-correct output, without learning much about morphology, while the encoder needs to learn quite a lot about source language morphology in order to create a good generic representation. However, their performance seems more or less comparable. 
In the following section we investigate what the role of the attention mechanism in the division of labor between encoder and decoder.

\begin{table}[t]
\centering
\begin{tabular}{c|rr|rr}
\toprule
& \multicolumn{2}{c|}{POS Accuracy} & \multicolumn{2}{c}{BLEU} \\
Attn & \texttt{ENC} & \texttt{DEC} & Ar-En &  En-Ar \\
\midrule
\cmark & 89.62 & 86.71 & 24.69 & 13.37 \\ 
\xmark & 74.10 & 85.54 & 11.88 & 5.04  \\ 
\bottomrule
\end{tabular}
\caption{POS tagging accuracy using encoder and decoder representations with/without attention.}
\label{tab:pos-dec-enc-attn-nogold}
\end{table}

\subsection{Effect of attention}

Consider the role of the attention mechanism in learning useful representations: during decoding, the attention weights are combined with the decoder's hidden states to generate the current translation. These two sources of information need to jointly point to the most relevant source word(s) and predict the next most likely word. Thus, the decoder puts significant emphasis on mapping back to the source sentence, which may come at the expense of obtaining a meaningful representation of the current word. We hypothesize that the attention mechanism might hurt 
the quality of the target word representations learned by the decoder. 

To test this hypothesis, we train NMT models with and without attention and compare the quality of their learned representations. 
As Table~\ref{tab:pos-dec-enc-attn-nogold} shows (compare 1st and 2nd rows), removing the attention mechanism decreases the quality of the encoder representations significantly, but only mildly hurts the quality of the decoder representations.  It seems that the decoder does not rely on the attention mechanism to obtain good target word representations, contrary to our hypothesis.

\subsection{Effect of word representation}

We also conducted experiments to verify our findings regarding word-based versus character-based representations on the decoder side. By character representation we mean a character CNN on the input words. The decoder predictions are still done at the word-level, which enables us to use its hidden states as word representations.

Table~\ref{tab:pos-dec-enc-word-char-nogold} shows POS accuracy of word-based %
vs.\ char-based representations in the encoder and decoder. 
In both bases, char-based representations perform better. 
BLEU scores behave differently: 
the char-based model leads to better translations in Arabic-to-English, but not in English-to-Arabic.
A possible explanation for this phenomenon %
is that the decoder's predictions are still done at word level even with the char-based model (which encodes the target input but not the output). In practice, this can lead to generating unknown words. Indeed, in 
Arabic-to-English 
the char-based model reduces the number of generated unknown words %
in the MT %
test set by 25\%, while in 
English-to-Arabic 
the number of unknown words %
remains roughly the same between word-based  %
and char-based models.

\section{Related Work} \label{sec:related-work}

\paragraph{Analysis of neural models}

The opacity of neural networks has motivated researchers to analyze such models in different ways. One line of work visualizes hidden unit activations in recurrent  neural networks 
that are trained for a given task \cite{elman1991distributed,karpathy2015visualizing,kadar2016representation,qian-qiu-huang:2016:EMNLP2016}. While such visualizations  
illuminate the inner workings of the network, 
they are often qualitative in nature and somewhat anecdotal. 
A different approach tries 
to provide a quantitative analysis by correlating parts of the neural  %
network with linguistic properties, for example by training a classifier to predict features of interest. Different units have been used, from word embeddings \cite{kohn:2015:EMNLP,qian-qiu-huang:2016:P16-11}, through LSTM gates or states \cite{qian-qiu-huang:2016:EMNLP2016}, to sentence embeddings \cite{adi2016fine}. Our work is most similar to \newcite{shi-padhi-knight:2016:EMNLP2016}, who use hidden vectors from 
a neural MT 
encoder to predict syntactic properties on the English source side. In contrast, we focus on representations in morphologically-rich languages and evaluate both source and target sides across several criteria.    
\newcite{vylomova2016word} also analyze different  %
representations for morphologically-rich languages in MT, but do not directly measure the quality of the learned representations.

\paragraph{Word representations in MT}
Machine translation systems that deal with morphologically-rich languages resort to various techniques for representing morphological knowledge, such as word segmentation \cite{C00-2162,E03-1076,Badr:2008:SES:1557690.1557732} 
and factored translation and reordering models \cite{koehn-hoang:2007:EMNLP-CoNLL2007,durrani-EtAl:2014:Coling}. Characters and other sub-word units have become increasingly popular in neural MT, although they had also been used in phrase-based MT for handling morphologically-rich \cite{Luong:D10-1015} or closely related language pairs \cite{durrani-EtAl:2010:ACL,Nakov:Tiedemann:2012}. In neural MT, such units are obtained in a pre-processing step---e.g.\ by byte-pair encoding \cite{sennrich-haddow-birch:2016:P16-12} or the word-piece model \cite{wu2016google}---or  learned during training 
using 
a character-based convolutional/recurrent sub-network \cite{costajussa-fonollosa:2016:P16-2,Luong:P16-1100,vylomova2016word}. The latter approach has the advantage of keeping 
the original word boundaries  without requiring pre- and post-processing. Here we focus on a character CNN which has been used in language modeling and machine translation \cite{kim2015character,belinkov-glass:2016:SeMaT,
costajussa-fonollosa:2016:P16-2,jozefowicz2016exploring,sajjad:2017:ACL}. We evaluate the quality of different representations learned by an MT system augmented with a character CNN in terms of POS and morphological tagging, and contrast them with a purely word-based system.

\begin{table}[t]
\centering
\begin{tabular}{l|rr|rr}
\toprule
& \multicolumn{2}{c|}{POS Accuracy} & \multicolumn{2}{c}{BLEU} \\
 & \texttt{ENC} & \texttt{DEC} & Ar-En &  En-Ar \\
\midrule
Word & 89.62 & 86.71 & 24.69 & 13.37 \\ 
Char & 95.35 & 91.11 & 28.42 & 13.00 \\ 
\bottomrule
\end{tabular}
\caption{POS tagging accuracy using word-based and char-based encoder/decoder representations.}
\label{tab:pos-dec-enc-word-char-nogold}
\end{table}

\section{Conclusion}

Neural networks have become ubiquitous in machine translation due to their elegant architecture and good performance. The representations they use for linguistic units are crucial for obtaining high-quality translation. In this work, 
we investigated how neural MT models learn word structure. We evaluated their representation quality on POS and morphological tagging in a number of languages. Our results lead to the following conclusions:
\begin{itemize}
\item Character-based representations are better than word-based ones for learning morphology, especially in rare and unseen words. 
\item Lower layers of the neural network are better at capturing morphology, while deeper networks improve translation performance. We hypothesize that lower layers are more focused on word structure, while higher ones are  focused on word meaning. 
\item  Translating into morphologically-poorer languages leads to better source-side
 representations. This is partly, but not completely,  %
 correlated with BLEU scores. %
\item There are only little differences between encoder and decoder representation quality. The attention mechanism does not seem to significantly affect the quality of the decoder representations, while it is important for the encoder representations. 
\end{itemize}

These insights can guide further development of neural MT systems. For instance, jointly learning translation and morphology can possibly lead to better representations and improved translation. Our analysis indicates that this kind of approach should take into account factors such as the encoding layer and the type of word representation. 

Another area for future work is to extend the analysis to other word %
representations (e.g.\ byte-pair encoding), 
deeper networks, 
and more 
semantically-oriented tasks such as 
semantic role-labeling or 
semantic parsing.

\section*{Acknowledgments}
We would like to thank Helmut Schmid for providing the Tiger corpus, members of the MIT Spoken Language Systems group for helpful comments, and the three anonymous reviewers for their useful suggestions. 
This research was carried out in collaboration between the HBKU Qatar Computing Research Institute (QCRI) and the MIT Computer Science and Artificial Intelligence Laboratory (CSAIL). 

\bibliography{acl2017}
\bibliographystyle{acl_natbib}

\newpage
\appendix

\section{Supplementary Material}
\label{sec:supplemental}

\subsection{Training Details} \label{sec:sup-training}

\paragraph{POS/Morphological classifier} The classifier used for all prediction tasks is a feed-forward network with one hidden layer, dropout ($\rho=0.5$), a ReLU non-linearity, and an output layer mapping to the tag set (followed by a Softmax). The size of the hidden layer is set to be identical to the size of the encoder's hidden state (typically 500 dimensions). We use Adam \cite{kingma2014adam} with default parameters to minimize the cross-entropy objective. 
Training is run with 
mini-batches of size 16 and stopped once the loss on the dev set stops improving; we allow a patience of 5 epochs.

\paragraph{Neural MT system} 
We train a 2-layer LSTM encoder-decoder with attention.  
We use the \texttt{seq2seq-attn} implementation \cite{kim2016} with the following default settings: word vectors and LSTM states have 500 dimensions, SGD with initial learning rate of 1.0 and rate decay of 0.5, and dropout rate of 0.3. The character-based model is a CNN with a highway network over characters \cite{kim2015character} with 1000 feature maps and a 
kernel width of 6 characters. This model was found to be 
useful for translating morphologically-rich languages 
\cite{costajussa-fonollosa:2016:P16-2}. The MT system is trained for 20 epochs, and the model with the best dev loss is 
used for extracting features for the classifier. 

\subsection{Data and Taggers} \label{sec:sup-data}
\paragraph{Datasets}
All of the translation models are trained on the Ted talks corpus included in WIT$^3$ \cite{cettoloEtAl:EAMT2012,cettolol:SeMaT:2016}. Statistics about each language pair are available on the WIT$^3$ website: \url{https://wit3.fbk.eu}.
For experiments using gold tags, we used the Arabic Treebank for Arabic (with the versions and splits described in the MADAMIRA manual \cite{PASHA14.593.L14-1479}) and the Tiger corpus for German.\footnote{\url{http://www.ims.uni-stuttgart.de/forschung/ressourcen/korpora/tiger.html}}

\paragraph{POS and morphological taggers}

We used the following tools to annotate the MT corpora: MADAMIRA \cite{PASHA14.593.L14-1479} for Arabic POS and morphological tags, Tree-Tagger \cite{schmid:2004:PAPERS} for Czech and French POS tags, LoPar \cite{schmid:00a} for German POS and morphological tags, and MXPOST \cite{ratnaparkhi98maximum} for English POS tags.
These tools are recommended 
on the Moses website.\footnote{\url{http://www.statmt.org/moses/?n=Moses.ExternalTools}} 
As mentioned before, our goal is not to achieve state-of-the-art 
results, but rather to study what different components of the NMT architecture learn about word morphology. 
Please refer to \newcite{mueller-schmid-schutze:2013:EMNLP} for representative POS and morphological tagging accuracies.

 \subsection{Supplementary Results}
 \label{sec:sup-results}

We report here results that were omitted from the paper due to the space limit. 
Table \ref{tab:different_layers} shows encoder results using 
different layers, 
languages, 
and  
representations (word/char-based). 
As noted in the paper,
all 
the results consistently show that i) layer 1 performs better than
layers 0 and 2;  
and
ii) char-based representations are better than word-based for learning morphology. 
Table \ref{tab:different_language} 
shows that translating into a morphologically-poor language (English) leads to better source 
representations, and Table \ref{tab:decoder} provides 
additional decoder results.

Table~\ref{tab:decoder-old} shows POS tagging accuracy using decoder representations, where the current word representation was used to predict the next word's tag. The idea is to evaluate whether the current word representation contains POS information about the output of the decoder. Clearly, the current word representation cannot be used to predict the next word's tag. This also holds when removing the attention (En-Ar, 85.54\%) or using character-based representations (En-Ar, 44.5\%). 
Since the decoder representation is in the pre-Softmax layer,  this means that most of the work of predicting the next work is done in the Softmax layer, while the pre-Softmax representation contains much information about the current input word. 

\newpage

 \begin{table}[h]
\centering
\begin{tabular}{l|r|r|r} 
\toprule
 & \multicolumn{1}{c}{Layer 0} & \multicolumn{1}{c}{Layer 1} & \multicolumn{1}{c}{Layer 2}\\ 
\midrule
& \multicolumn{3}{c}{Word/Char (POS)} \\ 
\midrule
De & 91.1/92.0 & 93.6/95.2 & 93.5/94.6 \\ 
Fr & 92.1/92.9 & 95.1/95.9 & 94.6/95.6 \\ 
Cz & 76.3/78.3 & 77.0/79.1 & 75.7/80.6 \\ 
\midrule
& \multicolumn{3}{c}{Word/Char (Morphology)}  \\ 
\midrule
De & 87.6/88.8 & 89.5/91.2 & 88.7/90.5 \\ 
\bottomrule
\end{tabular}
\caption{POS and morphology accuracy on  predicted tags using 
word- and char-based representations from 
different layers 
of *-to-En systems.}
\label{tab:different_layers}
\end{table}

 \begin{table}[h]
\centering
\begin{tabular}{l|r|r|r}
\toprule
\backslashbox{Source}{Target} & \multicolumn{1}{c}{English} & \multicolumn{1}{c}{Arabic} & \multicolumn{1}{c}{Self} \\
\midrule
German & 93.5 & 92.7 & 89.3 \\
Czech & 75.7 & 75.2 & 71.8  \\
\bottomrule
\end{tabular}
\caption{Impact of changing the target language on POS tagging accuracy. 
Self = German/Czech in rows 1/2 respectively.}
\label{tab:different_language}
\end{table}

\begin{table}[h]
\centering
\begin{tabular}{l|r|r|r|r}
\toprule
& En-De & En-Cz & De-En & Fr-En \\
\midrule
POS & 94.3 & 71.9 & 93.3 & 94.4 \\ 
BLEU & 23.4 & 13.9 & 29.6 & 37.8 \\
\bottomrule
\end{tabular}
\caption{POS accuracy and BLEU using decoder representations from different language pairs.}
\label{tab:decoder}
\end{table}

\begin{table}[h!]
\begin{tabular}{rrrrr}
\toprule
En-De & En-Cz & De-En & Fr-En & En-Ar \\
\midrule
53.6 & 36.3 & 53.3 & 54.1  & 43.9 \\
\bottomrule
\end{tabular}
\caption{Accuracy of predicting the next word's POS tag using decoder representations.}
\label{tab:decoder-old}
\end{table}

\end{document}